\documentclass[letterpaper, 10 pt, conference]{ieeeconf}  

\IEEEoverridecommandlockouts                              

\usepackage{multirow}
\usepackage{times}

\usepackage{hyperref}

\usepackage{color}
\usepackage{algorithm}
\usepackage{algpseudocode}
\usepackage{listings}
\usepackage{courier}
\usepackage{soul}
\usepackage{amsmath,amsfonts,amssymb}

\usepackage{mathtools}
\usepackage{subfigure}
\usepackage{wrapfig}
\usepackage{hyperref}
\usepackage{multirow}
\usepackage{graphicx}
\usepackage{graphicx}
\usepackage{soul}
\usepackage[flushleft]{threeparttable}
\usepackage{enumerate}
\algtext*{EndIf}
\algtext*{EndWhile}
\algtext*{EndFor}
\algtext*{EndProcedure}
\usepackage{algorithmicx}
\usepackage{algpseudocode}
\usepackage{epstopdf}

\definecolor{red}{rgb}{1.00,0.00,0.00}
\definecolor{blue}{rgb}{0.00,0.00,1.00}
\definecolor{green}{rgb}{0.2,0.70,0.2}
\definecolor{yellow}{rgb}{0.5,0.5,0.0}
\definecolor{white}{rgb}{1,1,1}

\title{\LARGE \bf
A Model-Based Balance Stabilization System for Biped Robot 
}

\author{Mohammadreza Kasaei$^{1}$, Nuno Lau$^{2}$ and Artur Pereira$^{2}$
\thanks{*This research is supported by Portuguese National Funds through Foundation for Science and Technology (FCT) through FCT scholarship SFRH/BD/118438/2016.}
\thanks{$^{1}$Mohammadreza Kasaei is with Department of Electronics, Telecommunications and Informatics~(IEETA/DETI) University of Aveiro, 3810-193 Aveiro, Portugal
        {\tt\small mohammadreza@ua.pt}}%
\thanks{$^{2}$Nuno Lau and Artur Pereira are with Faculty of the Electronics, Telecommunications and Informatics~(IEETA/DETI) University of Aveiro, 3810-193 Aveiro, Portugal
        {\tt\small \{nunolau, artur\}@ua.pt}%
}
}

\begin{document}

\maketitle
\thispagestyle{empty} 

\ieeefootline{Workshop on Modeling and Control of Dynamic Legged Locomotion: Insights from Template (Simplified) Models \\ IEEE/RSJ International Conference on Intelligent Robots and Systems (IROS) 2018, Madrid, Spain}


\begin{abstract}

This paper presents a model-based balance stabilization system which takes into account not only the stable part of COM dynamics but also the unstable part. In this system, the overall dynamics of a humanoid robot is approximated using a Linear Inverted Pendulum Plus Flywheel Model (LIPPFM). Moreover, Divergent Component of Motion~(DCM) is used to define when and where a robot should take a step to prevent falling. The proposed system has been successfully tested by performing several simulations using MATLAB. The simulation results show this system is capable of stabilizing the balance of the robot in various conditions.
\end{abstract}

\section{INTRODUCTION}
Humanoid robots are expected to take part in our daily life and perform different tasks in our real dynamic environments in a safe manner. In order to do that, keeping the stability in facing with uncertainties is the most critical requirement. 
During recent years, different types of controllers have been proposed which generally employed a simple dynamics model of the system to reduce the complexity of the system. Linear Inverted Pendulum Model~(LIPM)~\cite{kajita1991study} is one of the common successful models to develop dynamic walking for humanoid robots. Using this model, Kajita et al.~\cite{kajita2003biped} proposed a COM trajectory generator based on preview control and a predefined trajectory of the Zero Moment Point~(ZMP).
Since LIPM  does not consider the momentum around the COM and in other point of view, the body of a humanoid robot consists of several parts, their motions cause to generate momentum around the COM. To consider this momentum, Pratt et. al.~\cite{pratt2006capture} proposed an extended version of LIPM. This model tries to simulate the momentum around the COM by using a flywheel with a centroidal moment of inertia and rotational angle limits instead of single mass. Later, \cite{stephens2007humanoid} and \cite{kasaei2017reliable} used this model to determine decision surfaces that describe when and which particular strategies (e.g. ankle, hip or step) should be used to regain balance in the case of facing disturbances. In this paper, we extend our previous work~\cite{kasaei2018optimal} by measuring Divergent Component of Motion~(DCM) in each time step. Indeed, DCM is used to control the unstable part of COM without any effect on the stable part. 
\begin{figure}[!t]
	\label {DynamicModel}
	\centering
	\includegraphics[scale=0.25, trim= 2cm 0.5cm 2cm 0.0cm,clip]{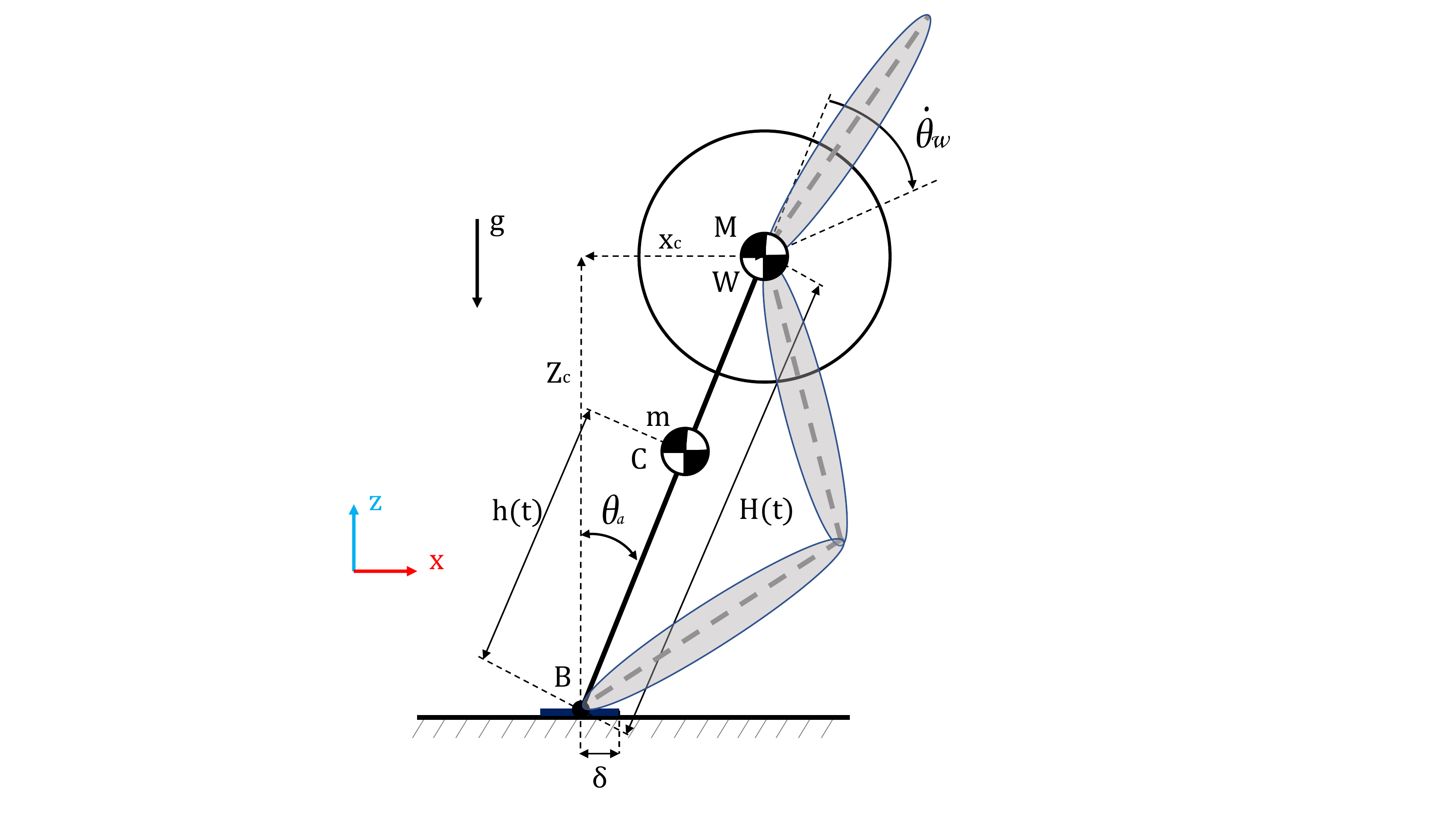}
	\vspace{-3mm}
	\caption{ Linear Inverted Pendulum Plus Flywheel Model. }
	\label{fig:DynamicModel}
		\vspace{-5mm}
\end{figure}
\vspace{-0mm}
\section{SYSTEM MODELING}
\label{sec:dynamics}
For investigating balance of a humanoid robot, LIPPFM model is used which is depicted in Fig.~\ref{fig:DynamicModel}. In this model, motions in X and Y direction are equivalent and independent. Therefore, to achieve simplicity, equations drive only for the X-direction. 
\vspace{-0mm}
\subsection{Motion Equations of Dynamics Model}
Euler-Lagrange equation is used to define the motion equations of LIPPFM:
%

\begin{equation}
\begin{aligned}
	\begin{bmatrix} \tau_a \\ \tau_w \end{bmatrix}  =&
	\begin{bmatrix} \gamma+I_{w} & I_{w} \\ I_{w} & I_{w} \end{bmatrix}
	\begin{bmatrix} \ddot{\theta_a}  \\ \ddot{\theta_w} \end{bmatrix} -
	\begin{bmatrix} \mu (g+\ddot{Z}_c) \sin\theta_a  \\ 0 \end{bmatrix} ,\\
	&
	\begin{cases}
		\gamma = M\times H^2 + I_p \\
		 \mu =  m\times h + M \times H
	\end{cases},
\end{aligned}
\label{eq:Lagrange2}
\end{equation}
\noindent
$M$ and $m$ are the masses of flywheel and the pendulum, $H$ and $h$ are the lengths from the base of the pendulum to flywheel center of mass and to pendulum center of mass respectively, $g$ describes the gravity acceleration, $\ddot{Z}_c$ represents the acceleration of CoM in Z-direction, $I_p$ is rotational inertia of pendulum about the base of pendulum and $I_w$ represents rotational inertia of flywheel around  flywheel center of mass.
By linearizing the model about the vertically upward equilibrium, $\theta_a = 0$, as well as by defining a new state vector, $x = [\theta_a \quad \dot{\theta_a} \quad \dot{\theta_w}]^\intercal$, the order of model will be reduced. Thus, by substituting $x$ into Equation~\ref{eq:Lagrange2} and rearranging in terms of $\dot{x}$, the system can be expressed as follows:
\begin{equation}
\begin{aligned}
\resizebox{\hsize}{!}{$
\begin{bmatrix} \dot{\theta}_a\\ \ddot{\theta}_a \\ \ddot{\theta}_w \end{bmatrix}
	=
	\begin{bmatrix} 
		0 & 1 & 0  \\ 
		\frac{\mu\times (g+\ddot{Z}_c)}{\gamma} & 0 & 0  \\
		\frac{-\mu\times (g+\ddot{Z}_c)}{\gamma} & 0 & 0  
	\end{bmatrix}	
	\begin{bmatrix} \theta_a\\ \dot{\theta_a} \\ \dot\theta_w \end{bmatrix}		
	+
	\begin{bmatrix} 
		0 & 0  \\
		\frac{1}{\gamma} & \frac{-1}{\gamma}  \\
		\frac{-1}{\gamma} & \frac{\gamma + I_w}{\gamma\times I_w}  
	\end{bmatrix}	
	\begin{bmatrix} 
		\tau_a  \\ \tau_w  
	\end{bmatrix}.$}
\end{aligned}	
\label{eq:statespace_alpha}
\end{equation}
\subsection{Reference Trajectory Tracking}
Equation~\ref{eq:statespace_alpha} represents our dynamics system in state space form which is a controllable system. Hence, a feedback controller for the system can be designed. 
In this paper, a Linear Quadratic Regulator~(LQR) is designed to track the desired trajectories and stabilize the system.


\subsection{Divergent Component of Motion Controller}
In case of larger disturbances, ZMP reaches to the border of support polygon and the robot is starting to roll over, therefore, neither ankle strategy nor hip strategy can keep the stability of robot. In this situation, humans often modify their footsteps and take a step to prevent falling. Several previous works \cite{pratt2006capture,englsberger2013three,hopkins2014humanoid} decompose the COM dynamics into stable and unstable components. The dynamics of the unstable part is generally used to define a 2D point~(Capture Point) or 3D point~(DCM) on which the robot should step to come to rest. This point can be defined as follow:

\begin{equation}
\zeta_x = x_c + \frac{\dot{x}_c}{\omega}
\label{eq:zeta}
\end{equation}
\noindent
where $\zeta_x$, $x_c$ represents position of DCM and COM in X-direction respectively, $\omega^2=\frac{g+\ddot{z}_c}{(z_c)}$ is the natural frequency of inverted pendulum. As is shown in Fig.\ref{fig:DynamicModel}, $x_c$ is equal to $ H\times\sin\theta_a$, if $\theta_a$ is considered to be small, $x_c$ can be approximated by $H\times\theta_a$. Now, by substituting $x_c=H\times\theta_a$ into Equation~\ref{eq:zeta} and by taking derivative from both sides, then by substituting $\ddot{\theta}_a$ form Equation~\ref{eq:Lagrange2} into this equation, $\dot{\zeta}_x$ will obtain as follow: 
\begin{equation}
\dot{\zeta}_x = H\times\big( \dot{\theta}_a + \frac{\tau_a - \tau_w+\mu(g+\ddot{Z}_c )\times\theta_a}{\omega\times\gamma} \big)
\label{eq:zetadot}
\end{equation}

According to this equation and the proposed dynamics system (Equation~\ref{eq:statespace_alpha}) , DCM can be measured using measurement equation of this system at each time step as follow: 
\begin{equation}
\begin{aligned}
y = \begin{bmatrix} \zeta_{mes} \\ \dot{\zeta}_{mes} \end{bmatrix} = C&x+Du \\
C = \begin{bmatrix} H & \frac{H}{\omega}&  0 \\
H & \frac{H \times\mu(g+\ddot{Z}_c)}{\omega\times\gamma}&  0  \end{bmatrix} &, D = \begin{bmatrix} 0&0 \\ \frac{H}{\omega\times\gamma} & \frac{-H}{\omega\times\gamma}\end{bmatrix}
\end{aligned}
\label{eq:measurement}
\end{equation}
\noindent
where C and D represent the output and the feedthrough matrix, respectively. Using this measurement, a PD controller can be applied to the error of DCM to modify the swing foot trajectory. 
\vspace{-0mm}
\section{SIMULATION RESULTS}
\label{sec:Results}
\begin{figure}[!t]
	\centering
	\begin{tabular}	{c}		
		\includegraphics[width=0.9\linewidth, trim= 1.5cm 0cm 2.5cm 0cm,clip]{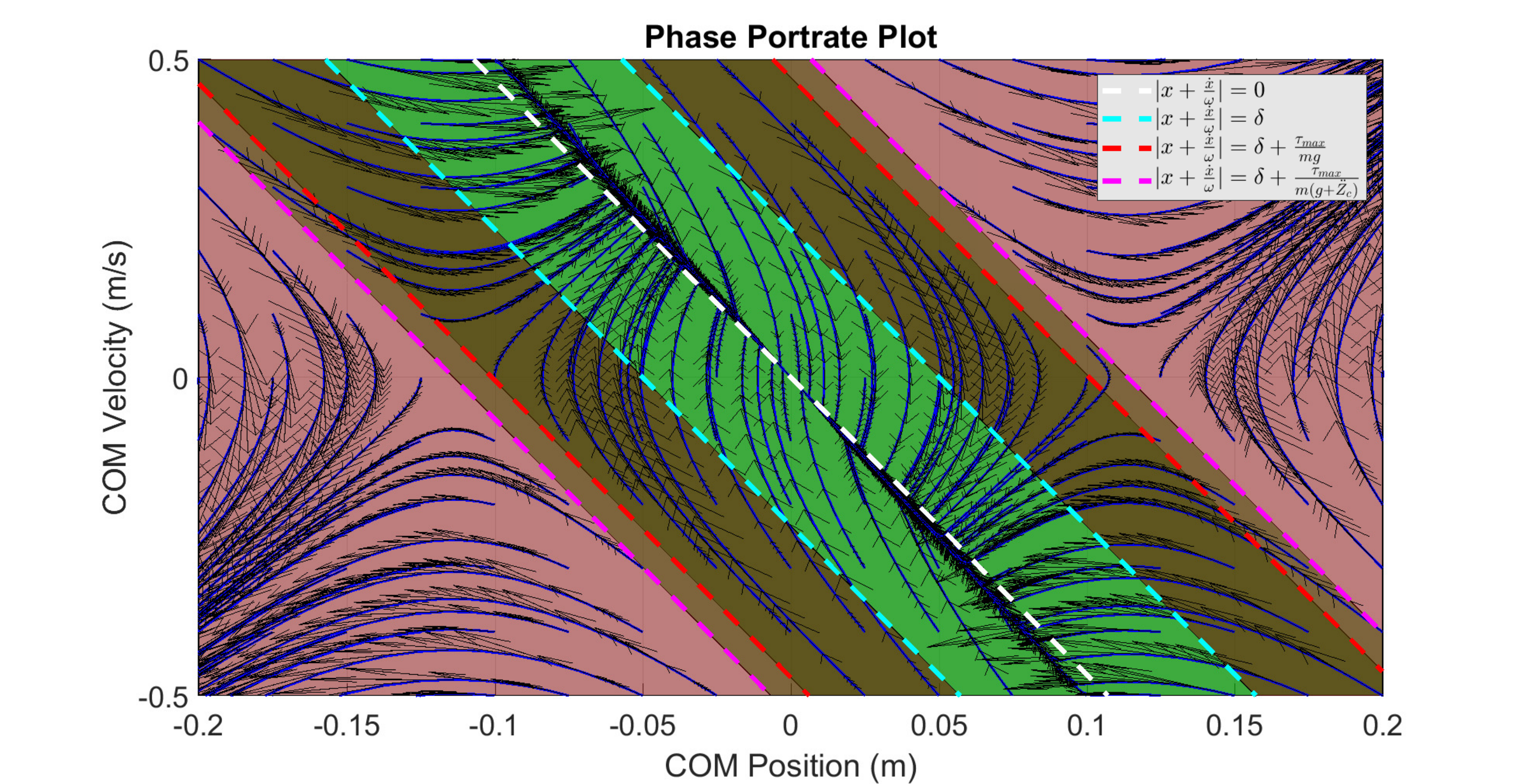}  
	\end{tabular}
	\vspace{-1mm}
	\caption{ Push recovery evaluation results. Pink areas represent unstable regions which robot should take a step to prevent falling. In other areas, the robot can keep its stability by using ankle or~(and) hip strategy.}
	\vspace{-4.5mm}
	\label{fig:push}
\end{figure}

The performance of the proposed controller verified via a simulation scenario which is performed using MATLAB. In this scenario, robot is considered to be in single support phase.
 The goal of this simulation is showing the robustness of the proposed controller in the presence of disturbances. To achieve this goal, in each trials, the simulated robot is started from different initial conditions. 
 The simulation results are shown in Fig.~\ref{fig:push}. As is shown in this figure, while the control signal ($u^*$) is not saturated, the controller is able to keep robot's stability, otherwise, the DCM controller should modify the landing position of swing foot.

%

\section{CONCLUSION}
\label{sec:Conclusion}

This paper proposed a model-based balance stabilization system which was composed of two controllers. The first controller was used to track reference trajectories by applying compensating torque at ankle and (or) hip joints. The outputs of this controller was used to measure DCM at each time step,  based on this measurement, another controller was designed and used to modify the swing foot trajectory. The performance of this system was verified using a simulation scenario. Simulation results show that system is able to stabilize the stability of the robot.


%

\bibliographystyle{IEEEtran}
\bibliography{WSIros2018}

\end{document}